\documentclass[robotics,article,accept,pdftex,moreauthors]{Definitions/mdpi} 
\newcolumntype{L}[1]{>{\raggedright\arraybackslash}m{#1}}
\newcolumntype{C}[1]{>{\centering\arraybackslash}m{#1}}
\newcolumntype{R}[1]{>{\raggedleft\arraybackslash}m{#1}}
\firstpage{1} 
\pubvolume{12}
\issuenum{4}
\articlenumber{114}
\pubyear{2023}
\copyrightyear{2023}
\externaleditor{Academic Editor: Donald Sofge}
\datereceived{19 June 2023} 
\daterevised{2 August 2023} 
\dateaccepted{5 August 2023 } 
\datepublished{7 August 2023 } 
\hreflink{https://doi.org/
10.3390/robotics12040114} 

\usepackage{amsmath}
\usepackage{multirow}

\usepackage{float}

\usepackage{scalerel}
\def\mylogo{\scalerel*{\includegraphics{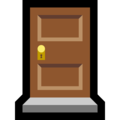}}{X}}

\Title{\mylogo{}SceneGATE: Scene-Graph Based Co-Attention Networks for Text Visual Question Answering}

\TitleCitation{SceneGATE: Scene-Graph Based Co-Attention Networks for Text Visual Question Answering}

\Author{{Feiqi} Cao $^{1,\dagger}$\orcidA{}, Siwen Luo $^{1,\dagger}$\orcidB{}, Felipe Nunez $^{1}$, Zean Wen $^{1}$, Josiah Poon $^{1}$\orcidC{} and Soyeon Caren Han $^{1,2,}$*\orcidD{}}

\AuthorNames{Feiqi Cao, Siwen Luo, Felipe Nunez, Zean Wen, Josiah Poon and Soyeon Caren Han}

\AuthorCitation{Cao, F.; Luo, S.; Nunez, F.; Wen, Z.; Poon, J.; Han, S.C.}

\address{%
$^{1}$ \quad {School of Computer Science, Faculty of Engineering, University of Sydney, \mbox{Camperdown, NSW 2006, Australia;}} {fcao0492@uni.sydney.edu.au (F.C.); siwen.luo@sydney.edu.au (S.L.); fnun9743@uni.sydney.edu.au~(F.N.); zwen2780@uni.sydney.edu.au~(Z.W.); josiah.poon@sydney.edu.au (J.P.)
}\\

$^{2}$ \quad {Department of Computer Science and Software Engineering, School of Physics, Maths and Computing,} {University} of Western Australia, {Crawley, WA 6009, Australia }}

\corres{Correspondence: {caren.han@uwa.edu.au}}

\firstnote{These authors contributed equally to this work.}

\abstract{Visual Question Answering (VQA) models fail catastrophically on questions related to the reading of text-carrying images. However, TextVQA aims to answer questions by understanding the scene texts in an image--question context, such as the brand name of a product or the time on a clock from an image. Most TextVQA approaches focus on objects and scene text detection, which are then integrated with the words in a question by a simple transformer encoder. The focus of these approaches is to use shared weights during the training of a multi-modal dataset, but it fails to capture the semantic relations between an image and a question. In this paper, we proposed a Scene Graph-Based Co-Attention Network (SceneGATE) for TextVQA, which reveals the semantic relations among the objects, the Optical Character Recognition (OCR) tokens and the question words. It is achieved by a TextVQA-based scene graph that discovers the underlying semantics of an image. We create a guided-attention module to capture the intra-modal interplay between the language and the vision as a guidance for inter-modal interactions. To permit explicit teaching of the relations between the two modalities, we propose and integrate two attention modules, namely a scene graph-based semantic relation-aware attention and a positional relation-aware attention. We conduct extensive experiments on two widely used benchmark datasets, Text-VQA and ST-VQA. It is shown that our SceneGATE method outperforms existing ones because of the scene graph and its attention modules.}

\keyword{artificial neural networks; computational and artificial intelligence; natural language processing; Visual Question Answering; scene graphs} 

\begin{document}

\section{Introduction}
Significant progress for multi-modal tasks that demand the simultaneous processing of both images and texts has been made in the past few years, and Visual Question Answering (VQA) is one of the prominent multi-modal tasks that requires answering natural language questions by inferring from the content of the given images. However, the nature of the questions and images of many existing VQA datasets is deficient in training the model to build a comprehensive understanding of human everyday scenes. For example, the collected photo-realistic images of many conventional VQA datasets exclude the texts that commonly appear in daily-life scenes, and the questions are merely designed to examine the recognition of objects and their attributes, such as colors and sizes. To overcome this limitation of existing conventional VQA datasets and to train models with better understandings of texts in realistic scenes through question answering, a new variant of VQA tasks, TextVQA~\cite{singh2019towards}, was recently proposed.

\begin{figure}[H]
    \centering
    \includegraphics[width=\textwidth]{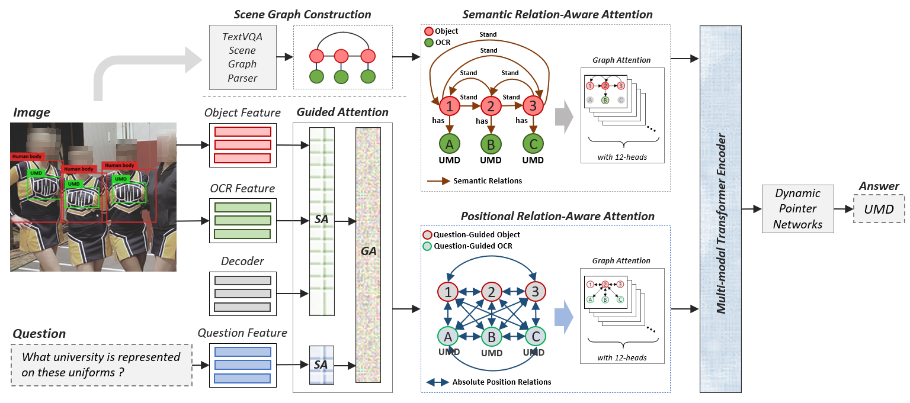}
    \caption{{Architecture} 
 of our SceneGATE model.}
    \label{fig:SceneGATE-archi}
\end{figure}

Images in TextQA tasks are collected from realistic scenes that contain various formats of texts, e.g., brand names and price tags, and the questions are specifically designed to be answered by referring to the textual information in the images. Hence, in addition to the recognition of objects as in the conventional VQA task, it is necessary for TextVQA models to additionally recognize the texts that are associated with the objects and capture these textual features from the images. Most current TextVQA models rely on the Optical Character Recognition (OCR) technique to directly extract the textual characters from images as OCR tokens and then integrate these OCR token features with image and question features for answer prediction, as shown in~\cite{singh2019towards, hu2020iterative,Zhu_Gao_Wang_Wu_2021}. However, the use of OCR tokens as an additional sequence of inputs to image object features and question word features can hardly reveal or capture the relations between the texts and their related objects in images. Such relations are significant in answering questions that show the explicit positional or semantic relationships between the objects and the textual characters, such as the relation {\textit{on}} 
 between \textit{UMD} and \textit{uniforms} in the question \textit{``What university is represented {\underline{on}} 
 these uniforms?''} in Figure~\ref{fig:SceneGATE-archi}. To capture the relatedness between objects and OCR tokens, some recent works~\cite{gao2020multi,kant2020spatially,han2020finding,gao2020structured} proposed to implicitly represent the relationships between objects and OCR tokens through their absolute locations. However, assuming such positional proximity as ``relatedness'' is not reliable and could be ambiguous, because irrelevant objects that are in different categories might be located in similar positions around one OCR token. 

To overcome this problem, we propose an explicit representation of the relationships between the OCR tokens and their associated objects in an image with the help of a scene graph. A scene graph~\cite{johnson2018image} is a graph structure that annotates the attributes of objects and the relationships between objects in an image. In this work, we propose a novel scene graph structure, specifically for TextVQA tasks, by assigning the OCR tokens as the attributes of objects to represent the affiliations of OCR tokens with their related objects. The scene graph embedding is encoded via the semantic embeddings of the objects and OCR tokens of the scene graph; thus, compared to previous works that only considered the visual features of objects, such scene graph embedding captures the relationships between objects and OCR tokens from the semantic aspect. A semantic relation-aware attention module is also applied to obtain the ultimate scene graph embedding that encodes the different semantic relationships between the objects and the OCR tokens. 

Another problem that this work tries to solve is to achieve much more intense interactions between the multi-modal inputs for TextVQA models and thus generate a better answer representation for the final answer prediction. Most of the previous works~\cite{hu2020iterative,gao2020structured} directly input different modalities (i.e., image, question and OCR token features) into a multi-modal transformer; although some works~\cite{han2020finding,Zhu_Gao_Wang_Wu_2021} made additional interactions for some modalities beforehand, these interactions were rather weak because there was a lack of simultaneous self-attention learning within each modality. Thus, we propose a Scene Graph-Based Co-Attention Network (SceneGATE) that includes a co-attention module consisting of two attention units: self-attention that boosts the intra-interactions for each modality and the guide-attention unit that uses the question features to guide the attention learning for the image and OCR token features. A positional relation-aware attention module is applied to the integrated visual features of objects and OCR tokens. Such integration enables the model to additionally learn relationships between the OCR tokens and their related objects from the positional level. The SceneGATE network operates on the two branches of scene graph features and the visual-level integrated multi-modal features in parallel, such that the relationships between objects and OCR tokens from both the semantic level and positional level are highlighted. The overall architecture of SceneGATE can be found in Figure ~\ref{fig:SceneGATE-archi}. In summary, the main contributions of our work are as follows.
\begin{itemize}
    \item To the best of our knowledge, this is the first attempt to apply a scene graph as an image representation in TextVQA. We introduce a novel scene graph generation framework for the TextVQA environment.
    \item We propose and integrate scene graph-based semantic relation-aware attention with positional relation-aware attention to establish a complete interaction between each question word and visual feature.
\end{itemize}

\section{Related Work}

\subsection{TextVQA}
\textit{Look, Read, Reason and Answer} (LoRRA)~\cite{singh2019towards} was the first baseline model for TextVQA tasks. It simply encodes the OCR tokens by FastText embeddings and enables the use of any type of contextual attention mechanism to integrate the questions, images and OCR token features. Further works have been proposed to solve the TextVQA problem from various aspects; for example, ref. \cite{hu2020iterative} proposed \textit{Multi-Modal Multi-Copy Mesh} (M4C) with an enriched OCR representation to capture more properties of the OCR tokens. Such enriched OCR token features are then projected with the object regions and question features into the same joint embedding space through the multi-modal transformer encoder. As an extension of M4C, ref. \cite{han2020finding} captured the relationships between the object regions and their related OCR tokens by encoding the OCR-related object features through softmax attention, which are the result of learning the locations of corresponding OCR tokens and the object regions. Others have proposed graph structures to better represent and encode the relationships between OCR tokens and the related object regions. For example, \mbox{ref. \cite{gao2020multi}} built three different graphs to represent the visual, semantic and numeric information of OCR tokens and object regions, where the nodes of the graphs are updated from locationally related nodes. Ref. \cite{kant2020spatially} proposed a spatially-aware graph attention module to emphasize the importance of different types of relative spatial relationships between OCR tokens and objects. Similarly, ref. \cite{gao2020structured} also utilized the graph attention module but this was conditioned on the relationships between the OCR tokens and the objects that were revealed from the question structural patterns. Ref. \cite{lu2021localize} proposed to align the related OCR tokens with the question by using a two-stage module that aligns the question with images via the pre-trained visual grounding task and then aligns the question and OCR tokens through object labels. Different from using an off-the-shelf OCR module, as in all the previous works, a recent~study \cite{zeng2021beyond} integrated the training of the OCR technique into the flow of an end-to-end TextVQA model to mitigate the influence of poor OCR accuracy on the final answer prediction. {A comparison of the TextVQA models can be found in Table~\ref{tab:textvqa-summary}.}

\subsection{Conventional VQA}
VQA~\cite{antol2015vqa} is a multi-modal task that requires answering natural language questions by looking at given images. The answer representation is generated from the inputs with different natures in essence: images formed by pixels and questions formed by semantic words. To align the image features and question features into the same joint embedding space for answer prediction, previous VQA works mainly adopt two different methods: fusion techniques \cite{shrestha2019answer, ben2019block, cadene2019murel, urooj2020mmft, han2021focal} and attention mechanisms. The attention mechanism in VQA tasks ranges from the basic vanilla attention \cite{hudson2018compositional, anderson2018bottom} to the recent, commonly used co-attention mechanism \cite{yu2019deep,gao2019multi,nguyen2018improved, rahman2021improved}. Such a co-attention mechanism aims to obtain a stronger interaction between different modalities by using the attention weights learned from individual modalities to guide the attention of each other. In this work, we use the guided-attention module for stronger intra-modal integration for OCR tokens, images and question features.

\begin{table}[H]
\caption{{Summary} of TextVQA models discussed.}\label{tab:textvqa-summary}
   \begin{adjustwidth}{-\extralength}{0cm}
 {
    \begin{tabular}{L{2.7cm}L{1.8cm}L{5cm}L{5cm}L{1.8cm}}
         \midrule
         \textbf{Model} & 
         \begin{tabular}[c]{@{}l@{}}\textbf{Question}\\ \textbf{Feature}\end{tabular} & 
         \begin{tabular}[c]{@{}l@{}}\textbf{Image} \textbf{Feature}\end{tabular} & 
         \textbf{Main Model} & 
         \textbf{Dataset} \\
         \midrule
         
         \begin{tabular}[c]{@{}l@{}}LoRAA  \cite{singh2019towards}\end{tabular} & 
         \begin{tabular}[c]{@{}l@{}}GloVe \\ LSTM \end{tabular} & 
         \begin{tabular}[c]{@{}l@{}}(Object) Faster R-CNN,\\ (OCR) FastText\end{tabular} & 
         \begin{tabular}[c]{@{}l@{}}Self-Attention, \\ OCR Copy Module \end{tabular} & 
         \begin{tabular}[c]{@{}l@{}}TextVQA,\\ VQA 2.0,\\ VizWiz\end{tabular} \\
         \midrule
         
         \begin{tabular}[c]{@{}l@{}}MM-GNN  \cite{gao2020multi}\end{tabular} & 
         LSTM & 
         \begin{tabular}[c]{@{}l@{}}(Object) Faster R-CNN,\\ (OCR) FastText/Sigmoid/\\ Cosine\end{tabular} & 
         \begin{tabular}[c]{@{}l@{}}Graph Neural Network,\\ Graph Attention, \\OCR Copy Module\end{tabular} & 
         \begin{tabular}[c]{@{}l@{}}TextVQA,\\ ST-VQA\end{tabular} \\
         \midrule
         
         \begin{tabular}[c]{@{}l@{}}M4C \cite{hu2020iterative}\end{tabular} & 
         BERT & 
         \begin{tabular}[c]{@{}l@{}}(Object) Faster R-CNN,\\ (OCR) Faster R-CNN +\\ FastText + PHOC\end{tabular} & 
         \begin{tabular}[c]{@{}l@{}}Multi-Modal Transformer,\\ Dynamic Pointer Network\end{tabular} & 
         \begin{tabular}[c]{@{}l@{}}TextVQA,\\ ST-VQA,\\ OCR-VQA\end{tabular} \\
         \midrule
         
         \begin{tabular}[c]{@{}l@{}}LaAP-Net \cite{han2020finding}\end{tabular} & 
         BERT & 
         \begin{tabular}[c]{@{}l@{}}(Object) Faster R-CNN,\\ (OCR) Faster R-CNN +\\ FastText + PHOC\end{tabular} & 
         \begin{tabular}[c]{@{}l@{}}Multi-Modal Transformer,\\ Localization Aware Predictor\end{tabular} & 
         \begin{tabular}[c]{@{}l@{}}TextVQA,\\ ST-VQA,\\ OCR-VQA\end{tabular} \\
         \midrule
         
         \begin{tabular}[c]{@{}l@{}}SMA \cite{gao2020structured}\end{tabular} & 
         BERT & 
         \begin{tabular}[c]{@{}l@{}}(Object) Faster R-CNN,\\ (OCR) Faster R-CNN +\\ FastText + PHOC\end{tabular} & 
         \begin{tabular}[c]{@{}l@{}}Self-Attention, \\Graph Atttention,\\ Dynamic Pointer Network\end{tabular} & 
         \begin{tabular}[c]{@{}l@{}}TextVQA,\\ ST-VQA\end{tabular} \\
         \midrule
         
         \begin{tabular}[c]{@{}l@{}}SA-M4C  \cite{kant2020spatially}\end{tabular} & 
         BERT & \begin{tabular}[c]{@{}l@{}}(Object) Faster R-CNN, \\ (OCR) Faster R-CNN +\\ FastText + PHOC\\\end{tabular} & 
         \begin{tabular}[c]{@{}l@{}}Multi-Modal Transformer,\\ Spatial Attention,\\ Dynamic Pointer Network\end{tabular} & 
         \begin{tabular}[c]{@{}l@{}}TextVQA,\\ ST-VQA\end{tabular} \\
         \midrule
         
         \begin{tabular}[c]{@{}l@{}}SNE \cite{Zhu_Gao_Wang_Wu_2021}\end{tabular} & 
          BERT& 
         \begin{tabular}[c]{@{}l@{}}(Object) Faster R-CNN, \\(OCR) Faster R-CNN +\\ FastText + PHOC +\\Recog-CNN\end{tabular} & 
         \begin{tabular}[c]{@{}l@{}}Feature Summarizing Attention, \\Multi-Modal Transformer,\\Dynamic Pointer Network\end{tabular} & 
         \begin{tabular}[c]{@{}l@{}}TextVQA,\\ ST-VQA\end{tabular} \\
         \midrule
    \end{tabular}}
\end{adjustwidth}
\end{table}

\subsection{Scene Graph in Visual Language Tasks}
The scene graph has been applied in various visual language tasks, including image captioning~\cite{yang2019auto, gu2019unpaired}, text to image generation~\cite{han2020victr} and image--text retrieval~\cite{wang2020cross}. Recently, scene graphs have also been used in VQA. For example, ref. \cite{luo2020rexup} processed the scene graphs and image features simultaneously through two parallel branches of recurrent memory networks to improve the model's reasoning ability over objects' relationships. \mbox{Ref. \cite{hudson2019learning}} proposed to use a probabilistic scene graph of images as the state machine, where questions were transformed into instructions to perform the reasoning process. Ref. \cite{haurilet2019s} claimed that only partial scene graphs are effective for answer prediction and proposed a selective system to choose the most important path in a scene graph and the most probable destination node on the graph to predict the answers. Ref. \cite{nuthalapati2021lightweight} used a graph attention network to encode scene graph embeddings to leverage the relatedness between different objects. Ref. \cite{wang2022sgeitl} applied the pre-training pipeline for Visual Commonsense Reasoning (VCR) tasks by incorporating object-based scene graphs in transformer layers to focus on semantically adjacent object nodes within multiple hops, regardless of relationship types. Nevertheless, our work is the first to apply a scene graph in TextVQA tasks and we propose a novel TextVQA-based scene graph structure to explicitly represent the affiliations between objects and OCR tokens.

\section{SceneGATE---Input Representations}\label{sec:sceneGATE_input}
In this work, we proposed a Scene Graph-Based Co-Attention Network for TextVQA tasks; the architecture is shown in Figure \ref{fig:SceneGATE-archi}. We first describe the input representations, the scene graph generation and the scene graph encoding methods in Section~\ref{sec:sceneGATE_input}. We then explain the co-attention networks for multi-modality semantic and positional relation integration in Section~\ref{sec:sceneGATE_atte}. {Table ~\ref{tab:symbol-def} defines all the mathematical symbols in our description.}

\vspace{-6pt}
\begin{table}[H]
\caption{{Symbols} and definitions.}\label{tab:symbol-def}
\centering
\small
 \scalebox{0.99}{\begin{tabular}{ll} 
 \toprule
 \textbf{Symbol} & \textbf{Definition} \\
 \midrule
    $\mathcal{SG} = (\mathcal{V}, \mathcal{E})$ & scene graph of an image with node set $\mathcal{V}$ and edge set $\mathcal{E}$ \\
    $sg_o \in O$ & object node set of each scene graph \\
    $sg_a \in A$ & attribute node set of each scene graph \\
    $sg_r \in R$ & relationship node set of each scene graph \\
    $P = |O \cup A|$ & total number of object and attribute nodes in each scene graph \\
    $N = |O|$ & total number of object nodes in each scene graph \\
    $M = |O \cup A \cup R|$ & total number of nodes in each scene graph \\
    $H^{(l)}$ & hidden states at layer $l$ of GCN \\
    $X_{GCN}$ & input scene graph node feature matrix of GCN \\
    $\tilde{A}$ & adjacency matrix of scene graph nodes \\
    $\tilde{D}$ & degree matrix of scene graph nodes \\
    ${w_1,\dots,w_t}$ & question words \\
    $d$ & embedding size of question words\\
    $d'$ & embedding size of OCR tokens \\
    \midrule
    $X$ & generalized input feature matrix into self-attention/guided attention \\
    $Y$ & generalized additional input feature matrix into guided attention \\
    $T$ & self-attended question representation \\
    $V = V_{obj}\cup V_{ocr} \cup D$ & self-attended visual objects, OCR features and decoder hidden states \\
    $V' = V_{obj}'\cup V_{ocr}' \cup D'$ & question-guided visual objects, OCR features and decoder hidden states \\
    $F_{s}$ & positional relation-aware (PRA) attention layer output \\
    $F_{sg}$ & semantic relation-aware (SRA) attention layer output \\
    $R_{j}$ & subset of relationships that the $j$-th head of SRA attention attends to \\
    $\kappa = |R_{j}|$ & number of relationships that each SRA/PRA head attends to \\
    $\beta$ & a bias term introduced in attention computation of SRA/PRA attention \\
    $t$ & time step at decoding stage\\
    $D_{(t)}'$, $D_{(<t)}'$, $D_{(>t)}'$ & decoder answer token at, before and after time step $t$, respectively \\
 \bottomrule
\end{tabular}}
\end{table}

\subsection{Input Representations}\label{sec:input_reprensentation}
Given question words ${w_1,\dots,w_t}$, we encode each word into a $d$-dimensional \textit{Bidirectional Encoder Representations from Transformers} (BERT) embedding~\cite{devlin-etal-2019-bert}. The weights are then fine-tuned during training. For objects in each image, we obtain the appearance features for a maximum of 100 objects via the Faster-RCNN model pre-trained on the Visual Genome dataset~\cite{Krishna2017VG}. We concatenate the appearance features of each object with its corresponding bounding box coordinates to represent each object region. For the OCR tokens, we extract their appearance features from the images using the same pre-trained Faster-RCNN model. Each OCR token {(Google} Cloud OCR Extractor: \url{https://cloud.google.com/products/ai/}{, accessed on 30 June 2023}) is also encoded by 300-dimensional pre-trained FastText embeddings {(FastText} embedding pre-trained with subword information on Wikipedia 2017, UMBC WebBase corpus and statmt.org news dataset: \url{https://fasttext.cc/docs/en/english-vectors.html}{, accessed on 30 June 2023})). Following \cite{hu2020iterative}, we concatenate the appearance features, FastText embeddings, Pyramidal Histogram of Characters (PHOC)~\cite{phoc} features and the bounding box coordinates of each OCR token to obtain the $d'$-dimensional enriched OCR embedding.

\subsection{Scene Graph Construction}\label{sec:sg-construction}
A scene graph is a graph structure $\mathcal{SG} = (\mathcal{V}, \mathcal{E})$ that denotes the relationships between objects as well as the associated attributes of each object for an image, where objects $sg_o \in O$, attributes $sg_a \in A$ and relationships $sg_r \in R$ are set as nodes of the graph. In this work, we construct a novel scene graph structure that is specific for TextVQA tasks to represent the affiliations between OCR tokens and the associated object regions. 

For each object in an image, we compare its bounding box coordinates with the other objects in the image. We have defined 11 different relation types: \textit{inside, surrounding, to the right of, to the left of, under, above, top right, bottom right, top left, bottom left} and {\em overlap}. {(The} semantic relation types can be defined in various ways if they can be represented in different semantic categories. This will be encoded in the categorical embedding in Section~\ref{sg_embedding}). We also compare the bounding box coordinates of each OCR token with all the objects in the image and assign this OCR token as the attribute of the object whose bounding box surrounds the OCR token's bounding box with the highest intersection over union (IoU) score. Finally, we obtain the triplet of $(sg_{o_{i}}, sg_r, sg_{o_{j}})$ for every two objects and the pair of $(sg_{o_{i}}, sg_{a_{i}})$ for each object that has attributes. We use bi-directional edges in the scene graph.

\subsection{Scene Graph Embedding}\label{sg_embedding}
We use different methods to encode the scene graph based on whether the (scene graph-based) semantic relation-aware (SRA) attention is applied or not. We describe the methods in detail in this section. We report our findings from the ablation studies in Section~\ref{sec:ablation}, where the SRA attention is not applied. 

When SRA attention is applied, we initialize the node features of each object and its attributes with a 300-dimensional embedding. They are stacked together as the node embedding matrix $Matrix^{P \times 300}$, where $P$ is the total number of objects and attribute nodes of each scene graph. We then add an extra relationship type, \textit{self}, to the 11 pre-defined relationship types as mentioned in Section~\ref{sec:sg-construction} in order to denote the relationship of each node with itself. In addition to the triplet of $(sg_{o_{i}}, sg_r, sg_{o_{j}})$ denoting the relationships between every two objects, we further add the relationships \textit{inside} and \textit{surrounding} between objects and their attributes to explicitly show and encode the existing semantic relationships between the objects and their associated OCR tokens. Hence, the object--attribute pair $(sg_{o_{i}}, sg_{a_{i}})$ now becomes a triplet of $(sg_{o_{i}}, sg_r, sg_{a_{i}})$, where $sg_r = \textit{surrounding}$, and $(sg_{a_{i}}, sg_r, sg_{o_{i}})$, where $sg_r$ would be \textit{inside}. The 12 relationship types are then converted into numeric labels of 1 to 12 to build the adjacency matrix that covers all the object nodes and attribute nodes of a scene~graph.

When SRA Attention is {\textbf{not}} applied, we encode each scene graph according to~\cite{luo2020rexup}. We first initialize the node embedding for all objects, relationship and attribute nodes in 300 dimensions, and then update each object node embedding with its associated attribute nodes' embeddings and all the relationship nodes' embeddings as well as their associated subject node embeddings. Specifically, for each $sg_{o_{i}}$, we update its embedding to a 900-dimensional embedding by concatenating two additional embeddings: (1) the average of all the related relationship embeddings, where each relationship embedding is the average of the embeddings of the relationship node and the subject node in the triplet of $(sg_{o_{i}}, sg_r, sg_{o_{j}})$ for $sg_{o_{i}}$; (2) the average of all the embeddings of the associated attribute nodes that are connected to $sg_{o_{i}}$. The updated object node embeddings are stacked into a matrix $Matrix^{N \times 900}$ as the scene graph representations for each scene graph, where $N$ is the number of total object nodes in each scene graph. We also propose two approaches to initialize the node features: pre-trained word embedding and GCN-based embedding, as illustrated in Sections~\ref{sec:node_ini_1} and~\ref{sec:node_ini_2}. The performance for these two approaches is compared in Section~\ref{sec:ablation}.

\subsubsection{Pre-Trained Word Embedding}\label{sec:node_ini_1}
Each node is initialized by either a 300-dimensional pre-trained GloVe embedding {(GloVe} embedding pre-trained on the Wikipedia and Gigaword5 corpus: \url{https://nlp.stanford.edu/projects/glove/}{, accessed on 30 June 2023}) or FastText embedding. For words with multiple tokens, we take the average of each token's embedding as the node embedding.

\subsubsection{GCN-Based Embedding}\label{sec:node_ini_2}
Graph convolutional networks (GCN) take the node embedding matrix and the adjacency matrix as inputs. They are propagated over all nodes and result in a matrix with the updated node features. We construct one graph based on all the unique categories of objects, relationships and attributes nodes across all the scene graphs of all the images in the dataset, and we propagate them over a 2-layer GCN for the updated node representations following Equation~\eqref{equation:gcn_layer}.
\begin{equation} \label{equation:gcn_layer}
    H^{(l+1)} = f\left ( \tilde{D}^{-\frac{1}{2}}\tilde{A}\tilde{D}^{-\frac{1}{2}}H^{(l)}W^{(l)} \right )
\end{equation}

$H^{(l)}$ at layer $l = 0$ is the input node feature matrix $X_{GCN}\in\mathbb{R}^{M\times M}$, where each node is represented by a one-hot encoding and $M$ is the total number of nodes. Based on the triplets of $(sg_{o_{i}}, sg_r, sg_{o_{j}})$ and the pairs of $(sg_{o_{i}}, sg_{a})$ in the scene graphs, the objects are connected with the related relations and the associated attributes in the adjacency matrix $\tilde{A}$. We assign a weight of 1 to all the connected edges and 0 to non-edges in $\tilde{A}$. $\tilde{D}$ is the degree matrix computed based on $\tilde{A}$ such that $\tilde{D}_{ii}=\sum_{j}{\tilde{A}_{ij}}$. We train two different GCNs in terms of different node labels. In the object-GCN, we manually categorize all the object types into 60 super-classes based on the hypernym of the synset (synonym set) of each object token and use these 60 super-classes as the labels of each object node during GCN training. In the attribute-GCN, we label each attribute node with the attribute tokens' named entity recognition (NER) types. {(We} used the Google Cloud NLP API to identity the entity type: \url{https://cloud.google.com/natural-language/docs/analyzing-entities}{, accessed on 30 June 2023}).
We have 8 different classes: CONSUMER GOOD, EVENT, LOCATION, NUMBER, ORGANIZATION, PERSON, WORK OF ART and OTHER. The outputs of the second layer of object-GCN $H_{obj}^{(l_2)}$ and attribute-GCN $H_{att}^{(l_2)}$ are then passed to a minimum pooling layer as in Equation~\eqref{equation:poling} to obtain the final node feature matrix $X_{GCN}^{'}\in\mathbb{R}^{M\times d}$.
\begin{equation} \label{equation:poling}
    Pooling_{min} = min(H_{obj}^{(l_2)}, H_{att}^{(l_2)})
\end{equation}

The node features of $X_{GCN}^{'}$ are then used as the initial node representations for object, relationship and attribute to generate the scene graph embedding of each scene graph.

\section{SceneGATE---Co-Attention Networks}\label{sec:sceneGATE_atte}
For multi-modality integration, we apply a guided attention module over the inputs and introduce two parallel branches of scene graph-based semantic relation-aware attention and positional relation-aware attention layers.

\subsection{Self-Attention Module}\label{sec:sa}
The self-attention (SA) module consists of a multi-head attention layer and a feed-forward layer with ReLU activation and dropout~\cite{vaswani2017tranformer}. The input matrix $X$ is transformed into three matrices that are in the same dimensions, i.e., query, key and value, as the learnable weights. These three matrices are then fed into a multi-head attention layer for the calculation of scaled dot-product attention. We respectively apply two SA modules for our two sets of inputs $X$: (1) the question features to obtain the self-attended question representations $T$; and (2) the combination of object appearance features, the enriched OCR token features as described in~Section~\ref{sec:input_reprensentation} and the answer token features from the decoder, to obtain the attended visual-level object-OCR features and decoder hidden states $V = V_{obj}\cup V_{ocr} \cup D$.

\subsection{Guided Attention Module}

The guided attention (GA) module shares the same structure and hyperparameters as the SA module, but the inputs of the multi-head attention layer are the feature matrix $X$ and the transformed key and value matrices of another feature matrix $Y$. In the GA module, we use the self-attended question representations $T$ as the feature matrix $Y$ to guide the attention learning with the attended visual-level object-OCR features $V$ that function as the input feature matrix $X$. Finally, we obtain the question-guided object-OCR features and decoder hidden states $V' = V_{obj}'\cup V_{ocr}' \cup D'$.

\subsection{Semantic Relation-Aware Attention}\label{sec:sra-attention}
We use the transformer encoder \cite{vaswani2017tranformer} with 12 heads as the backbone for our semantic relation-aware (SRA) attention Layer. As illustrated in~Section~\ref{sg_embedding}, we annotate 12 pre-defined different relationships $sg_r \in R$ between every two object nodes $(sg_{o_{i}}, sg_r, sg_{a_{i}})$ and 2 relationship types between objects and their attributes $(sg_{a_{i}}, sg_r, sg_{o_{i}})$ in the scene graph. We introduce a special attention mechanism whereby each head of the transformer will only attend to certain nodes of the scene graph. In other words, we only allow each node to attend to nodes that are connected by certain types of relationships $R_{j}$ for the $j$-th head in the SRA attention Layer, where $R_{j}$ is a subset of the 12 relationships $R$ and contains only $\kappa$ number of relationship types. A bias term $\beta$ is added to the calculation of the scaled dot-product attention. When $\beta = 0$, the attention weights can be calculated normally between the two nodes, considering that an edge is mapped to some relationship types $sg_{r} \in R_{j}$ that the $j$-th head of the SRA attention Layer is supposed to attend to. When $\beta = -\infty$, the attention weights between two nodes also become $-\infty$, considering that an edge is mapped to a set of relationship types $sg_{r} \notin R_{j}$. Since each head is supposed to attend to specific sub-information in a scene graph, the calculation of the attention is only limited to a given set of nodes. In order to manage which relationship types and the number of relationship types that each head in the SRA attention layer pays attention to, we need to control the values of $\beta$ and $\kappa$. Empirically, we find that $\kappa = 3$ is the most suitable.

\subsection{Positional Relation-Aware Attention}
Inspired by \cite{kant2020spatially}, we also construct a directed complete spatial graph over the object features $V_{obj}’$ and OCR token features $V_{ocr}’$ in $V’$, where each edge corresponds to one of the spatial relationship types according to their relative positions. Additional edges are also added to connect all the object nodes and OCR tokens to all the question tokens.

Similar to the SRA attention Layer, the positional relation-aware (PRA) attention layer also uses the structure of the multi-modal transformer encoder with 12 heads as the backbone and permits each head to attend to different subsets of the spatial relationship types. All the heads also allow all the objects and OCR tokens to attend to the question's~words. 

Moreover, a causal attention mask is applied for the decoder $D’$ in the PRA attention layer. $D_{(t)}'$ is the answer token generated from the decoder at time step $t$. The attention layer can attend to all question tokens, objects and OCR tokens along with the previously decoded entries in the answer $D_{(<t)}'$, without attending to $D_{(>t)}'$, the decoding entries after time step $t$. $T$ and $V'$, obtained from the SA module and GA module, are combined as the input sequence and fed to two subsequent PRA attention layers to obtain spatially attended features $F_{s}$.

Outputs $F_{sg}$ from the SRA attention layer are combined with $F_{s}$ to become the input to the multi-modal transformer encoders~\cite{vaswani2017tranformer}. The combined input allows the model to attend to all input features in a pair-wise manner. The multi-word answer in each time step $t$ is decoded using the dynamic pointer network following~\cite{hu2020iterative}.

\section{Experiments}

\subsection{Datasets}
We evaluate our model with two widely used benchmark datasets: Text-VQA and ST-VQA. The Text-VQA dataset was proposed by~\cite{singh2019towards} in 2019. Different from the conventional VQA datasets, images in the Text-VQA dataset contain texts in different formats, and the questions are specifically designed to be answered by referring to the textual information in images. The Text-VQA dataset collects 28,408 images that contain texts from the Open Images v3 dataset \cite{openimages}. There are 45,336 question--image pairs in the Text-VQA dataset, which are split into 34,602, 5000 and 5734 for training, validation and testing, respectively. Each question has 10 ground truth answers, and the voting of these 10 answers is used to compute the soft accuracy score. The ST-VQA dataset~\cite{biten2019scene} is a concurrent work of the Text-VQA dataset. However, different from the Text-VQA dataset, the 23,038 images in the ST-VQA dataset are collected from multiple source image datasets, including the Coco-text~\cite{veit2016cocotext}, Visual Genome~\cite{Krishna2017VG}, VizWiz~\cite{gurari2018vizwiz}, ICDAR~\cite{Karatzas2013ICDAR,Karatzas2015ICDAR}, ImageNet~\cite{deng2009imagenet} and IIIT-STR~\cite{mishra2013iiitstr} datasets, in order to reduce the effect of possible biases from a single-source image dataset. There are 17,028 images/23,446 questions for the training set, 1893 images/2628 questions for the validation set and 2971 images/4070 questions for the test set. Each question has at most 2 ground truth answers to compute the accuracy score by soft voting, similar to the VQA context. To clearly show the difference between conventional VQA datasets and the Text-VQA/ST-VQA dataset, we list some typical conventional VQA datasets and compare them in terms of various aspects in Table~\ref{tab:dataset_compare}.

\begin{table}[H]
\caption{{Comparison} between conventional VQA dataset and TextVQA/ST-VQA dataset.}\label{tab:dataset_compare}
\centering
	\begin{adjustwidth}{-\extralength}{0cm}
		\scalebox{1}{\begin{tabular}{L{2cm}C{1cm}C{4cm}C{4cm}C{3cm}C{2cm}}
                \toprule
                \textbf{Dataset} & \textbf{Year} & \textbf{Image Type} & \textbf{Question Type} & \textbf{Answer Type} & \textbf{Size (Img./Q.)} \\ 
                \midrule
                VQA v1.~\cite{antol2015vqa} & 2015 & photo-realistic images + abstract scene & asking different attributes of objects & open-ended/multiple choice & 255 K/760 K \\ \midrule
                VQA v2.~\cite{goyal2017making} & 2017 & photo-realistic images & asking different attributes of objects & open-ended & 204 K/1.1 M \\ \midrule
                Clevr~\cite{johnson2017clevr} & 2017 & auto-generated synthetic images & compositional questions & open-ended & 100 K/865 K \\\midrule
                Visual Genome~\cite{Krishna2017VG} & 2017 & photo-realistic images & asking different attributes of objects & open-ended & 108 K/1.7 M \\\midrule
                GQA~\cite{hudson2019gqa} & 2019 & photo-realistic images & compositional questions & open-ended & 113K/22M \\\midrule
                Text-VQA~\cite{singh2019towards} & 2019 &photo-realistic images contain texts & asking for texts in images & open-ended & 28 K/45 K \\\midrule
                ST-VQA~\cite{biten2019scene} & 2019 & photo-realistic images contain texts & asking for texts in images & open-ended & 23 K/30 K \\
                \bottomrule
            \end{tabular}}
	\end{adjustwidth}
\end{table}

\subsection{Implementation Details}
We encode the question features, the appearance features of objects and the OCR tokens in the same dimension and the same maximum sequence length as in~\cite{kant2020spatially}. Each scene graph has an average of 36 object nodes and a maximum of 100 OCR nodes. The downstream SA and GA modules have a dimension of 768, with 8 attention heads and a dropout rate of 0.1. Our experiments are conducted utilizing an NVIDIA Titan RTX GPU with 24 GB RAM, a 16 Intel(R) Core(TM) i9-9900X CPU @ 3.50 GHz with 128 GB RAM and the operating system of Ubuntu 20.04.1. Our final model contains around 95~million trainable parameters and requires around 0.6 h to train one epoch. The validation accuracy converges within 40 epochs of training for most of our model variants, and our best model converges within 8 epochs on both datasets. In short, the training of our best-performing model requires around 3GB GPU RAM and 4 h to complete. We use a batch size of 8~and follow the same settings as in~\cite{kant2020spatially} for other hyperparameter values. Details of all hyperparameters can be found in Appendix~\ref{appendix-hp}.

\subsection{Baseline Models}
We compare the SceneGATE network with the following baselines in this work: 
{\bf LoRRA}~\cite{singh2019towards} encodes the OCR tokens with only FastText embedding and it has an attention mechanism to integrate all image, question and OCR token features into the same joint embedding space for answer prediction. 
{\bf M4C}~\cite{hu2020iterative} uses enriched OCR token representations that include the appearance, semantic, character-level and spatial features of OCR tokens. The multi-modal transformer encoder is used for modality integration and iterative decoding, while a dynamic pointer network is applied for answer generation. 
\textit{Simple is not Easy} ({\bf SNE})~\cite{Zhu_Gao_Wang_Wu_2021} has three separate vanilla attention blocks for the independent integration of object region, OCR visual-based and OCR textual-based features with the questions.
\textit{Localization-Aware Answer Prediction} ({\bf LaAP-Net})~\cite{han2020finding} integrates objects and OCR token features via an attention mechanism to obtain the OCR-related image features, which is followed by M4C using the multi-modal transformer for integration with the question features. 
The \textit{Multi-Modal Graph Neural Network} ({\bf MM-GNN})~\cite{gao2020multi} constructs three graphs for object regions, semantic OCR tokens and numeric OCR tokens, which all interact to learn from the related nodes. 
\textit{Spatially Aware Multi-Modal Multi-Copy Mesh} ({\bf SA-M4C})~\cite{kant2020spatially} adopts a spatially aware self-attention module to capture and to encode 12 different types of spatial relationships between objects and OCR tokens.  
\textit{Structured Multi-Modal Attention} ({\bf SMA})~\cite{gao2020structured} applies a question-conditioned graph attention module to identify the potential relationships between objects and OCR tokens from the question patterns.

\section{Results}

\subsection{Performance Comparison}

Different from other works that used ST-VQA to enlarge the training dataset size, we compared the performance of our model with different baselines by training only on the original Text-VQA dataset. We can see from Table~\ref{tab:textvqa} that our model outperformed all the baselines and yielded the state-of-the-art result of 42.37\% validation accuracy and 44.02\% test accuracy on the Text-VQA dataset. We ran the code provided by SA-M4C to train the Text-VQA dataset with their default hyperparameters. Their results were only 40.71\% and 42.61\% for the validation and test accuracy, respectively, which were almost 2\% lower than our model. Compared with M4C, SNE and LoRRA, which simply used the attention mechanism to integrate the different modalities, the models (SMA, SA-M4C and SceneGATE) that applied the graph attention module achieved better performance, indicating the importance of the explicit representation and encoding of the relationships between objects and their related OCR tokens for the TextVQA task. However, both SMA and SA-M4C considered only the visual representation of object nodes and their spatial relationships, when the node embedding of OCR tokens in a graph was updated. Our model overcomes this limitation via the additional encoding of the semantic embeddings of object nodes and the semantic relationships among different objects and OCR tokens with the use of the scene graph. Our results also indicate the importance of such semantic relationship representation and the scene graph in the TextVQA task.

In addition to the accuracy rate, we also used the Average Normalized Levenshtein Similarity (ANLS) score, which was proposed for the evaluation of the ST-VQA dataset~\cite{biten2019scene}, as an additional evaluation metric to evaluate the performance on ST-VQA. The ANLS score aims to eliminate the dropped performance caused by OCR recognition errors. It compares the similarities between the ground truth answers and the prediction results, rather than the robust identity, as when using the accuracy rate. The edit distance of converting a prediction string into a ground truth string is measured by this metric in order to give a soft score for the prediction. If the edit distance is greater than 0.5, it can be considered as an incorrect prediction not resulting from OCR recognition mistakes, and a score of 0 is given. Otherwise, the difference of this edit distance from 1.0 is awarded as the prediction score. A higher ANLS score indicates that more accurate predictions are made by the model. The performance of our models and the baselines is compared in Table~\ref{tab:STVQA} and we can see that our model greatly outperforms the baselines by achieving 41.29\%, 0.525 and 0.516 for the validation accuracy, validation ANLS score and test ANLS score, respectively. Our model achieved a 1.5\% improvement in accuracy,  a 0.028 improvement for the validation ANLS score and a 0.039 improvement for the test ANLS score compared to the base model SA-M4C, which performed slightly worse than its base model M4C in terms of accuracy and achieved only around a 0.01 increase in the ANLS score on both the validation set and test set compared to its base model, M4C. The larger performance gap achieved by our model compared to our base model shows the importance of considering the semantic relations among the objects and OCR tokens.

\begin{table}[H]
\caption{{Results} on Text-VQA dataset. {Acc. refers to the soft accuracy score.}}\label{tab:textvqa}
 \setlength{\tabcolsep}{12.0mm}{\begin{tabular}{lcc} 
 \toprule
 \textbf{Model} & \textbf{{Acc.} on Val} & \textbf{Acc. on Test}\\
 \midrule
 LoRRA & 26.56 & 27.63\\
 MM-GNN & 32.92 & 32.46\\
 M4C & 39.40 & 39.01\\
 LaAP-Net & 40.68 & 40.54\\
 SMA & 40.05 & 40.66\\
 SA-M4C & 40.71 & 42.61\\
 SNE & 40.38 & 40.92\\
 \midrule
 {\textbf{SceneGATE (Ours)}} & {\textbf{42.37}} & {\textbf{44.02}}\\
 \bottomrule
\end{tabular}}
\end{table}
\unskip

\begin{table}[H]
\caption{{Results} on ST-VQA dataset. {Acc. refers to the soft accuracy score.}}\label{tab:STVQA}
\setlength{\tabcolsep}{9.7mm}{\begin{tabular}{lccc} 
 \toprule
 \multirow{2}{*}{\textbf{Model}} & \textbf{Acc.} & \textbf{ANLS} & \textbf{ANLS}\\
 & \textbf{on Val} & \textbf{on Val} & \textbf{on Test}\\
 \midrule
 MM-GNN & - & - & 0.207 \\
 M4C & 38.05 & 0.472 & 0.462 \\
 LaAP-Net & 39.74 & 0.497 & 0.485 \\
 SMA & - & - & 0.466 \\
 SA-M4C & 37.86 & 0.486 & 0.477 \\
 SNE & - & - & 0.509\\
 \midrule
 {\textbf{SceneGATE (Ours)}} & {\textbf{41.29}} & {\textbf{0.525}} & {\textbf{0.516}} \\
 \bottomrule
\end{tabular}}
\end{table}

\subsection{Ablation Studies}\label{sec:ablation}

\textbf{SG Embedding.} The use of a scene graph to capture the explicit semantic relationships between objects and OCR tokens makes an important contribution to our model's SOTA performance on both the Text-VQA and ST-VQA datasets. To examine the impact of different scene graph node embedding initialization methods on the model's performance, we also evaluated the model's performance in using GCN and GloVe for node embedding initialization on both the Text-VQA and ST-VQA validation sets. We can see from Table~\ref{tab:ablation1} that GloVe had the worst results, with only 41.33\% and 40.79\% accuracy on both the Text-VQA and ST-VQA validation sets, while the use of GCN would increase the performance slightly. FastText resulted in the best performance considering that FastText is more capable of dealing with the OOV issue for the cases of rare OCR tokens in a scene graph.

\textbf{Network Component.} To investigate the contribution of our model's components, we first integrated all the image, question, OCR token and scene graph features via multi-modal transformer encoders as in M4C~\cite{hu2020iterative}. This simple approach achieved 39.13\% and 37.78\% accuracy on the Text-VQA and ST-VQA validation sets, respectively, as shown in Table~\ref{tab:ablation2}. After the addition of the co-attention module for the better inter- and intra-integration of the image, question and OCR token features, the performance on the validation and test sets increased significantly to 41.27\% and 39.57\%, indicating the effectiveness of such self-attention-based guided attention. The inclusion of PRA attention layers gave an improvement in the accuracy rate by around 0.7\%, and the performance rose further to 42.37\% and 41.29\% for the Text-VQA and ST-VQA validation sets after adding the SRA attention layer over the scene graphs. These results prove the critical roles of the co-attention, PRA and SRA attention layers in our model.
\begin{table}[H]
\small
\caption{{Validation} performance of our model obtained on different types of scene graph node embedding. {Acc. refers to the soft accuracy score.}}\label{tab:ablation1}
\setlength{\tabcolsep}{13mm}{\begin{tabular}{lcc} 
 \toprule
 \multirow{2}{*}{\textbf{Model}} & \textbf{Acc. on} & \textbf{Acc. on}\\
 & \textbf{Text-VQA} & \textbf{ST-VQA}\\
 \midrule
 
 SceneGATE w GloVe & 41.33 & 40.79\\
 SceneGATE w GCN & 41.36 & 40.86\\
 {\textbf{SceneGATE w FastText}} & {\textbf{42.37}} & {\textbf{41.29}}\\
 \midrule
\end{tabular}}
\end{table}
\vspace{-6pt}

\begin{table}[H]
\small
\caption{{Ablation} testing results on the validation set. PRA: positional relation-aware. SRA: semantic relation-aware. {Acc. refers to the soft accuracy score.}}\label{tab:ablation2}
\setlength{\tabcolsep}{14mm}{\begin{tabular}{lcc} 
 \midrule
 \multirow{2}{*}{\textbf{Model}} & \textbf{Acc. on} & \textbf{Acc. on}\\
 & \textbf{Text-VQA} & \textbf{ST-VQA}\\
 \midrule
 MM Transformer & 39.13 & 37.78\\
 + Guided Attention & 41.27 & 39.57 \\
 + PRA Attention & 41.95 & 40.37\\
 {\textbf{+ SRA Attention}} & {\textbf{42.37}} & {\textbf{41.29}}\\
 \bottomrule
\end{tabular}}
\end{table}

\textbf{Effect of Layer Number.} In order to determine how many layers we should apply to each module of the model, we conducted experiments with different combinations of layer numbers and the results are presented in Table \ref{tab:hp-layerno}. Since having two multi-modal transformer encoder (MMTE) layers worked the best for the SA-M4C model \cite{kant2020spatially}, we started by fixing the number of final BERT layers to two and adopted all the combinations of numbers in the range [1, 3] for the number of PRA attention layers and the number of SRA attention Layers. We empirically observed that models with two PRA attention layers always performed better than others (rows 4--6 vs. other rows), yielding validation accuracy of more than 41.8\%. In addition, having two layers for each type of attention layer worked the best. Based on these two observations, we fixed the number of SPA layers to two and tested the model with a smaller number of SRA attention layers and MMTE layers. Eventually, we found that having two PRA attention layers, one SRA attention layer and one MMTE layer was able to yield the best validation accuracy result, 42.37\%.

\vspace{-6pt}
\begin{table}[H]
\small
\caption{{Validation} performance for different numbers of each type of attention layer used. MMTE: multi-modal transformer encoder layers, PRA: positional relation-aware attention layers, SRA: semantic relation-aware attention layers, {Acc.: soft accuracy score.}}
\label{tab:hp-layerno}
\setlength{\tabcolsep}{11.4mm} \begin{tabular}{c c c c} 
 \toprule
 \textbf{\# MMTE} & \textbf{\# PRA} & \textbf{\# SRA} & \textbf{Acc. on Val} \\
 \midrule
 2 & 1 & 1 & 41.33 \\
 2 & 1 & 2 & 41.73 \\
 2 & 1 & 3 & 41.24 \\
 2 & 2 & 1 & 41.82 \\
 \underline{2} & \underline{2} & \underline{2} & {\underline{41.89}} \\
 2 & 2 & 3 & 41.86 \\
 2 & 3 & 1 & 41.68 \\
 2 & 3 & 2 & 41.32 \\
 2 & 3 & 3 & 41.19 \\
 \midrule
 1 & 2 & 2 & 41.53 \\
 {\textbf{1}} & {\textbf{2}} & {\textbf{1}} & {\textbf{42.37}} \\
 \bottomrule
\end{tabular}
\end{table}

\subsection{Quality Analysis}

\begin{figure}[H]
    \includegraphics[width=8.5cm]{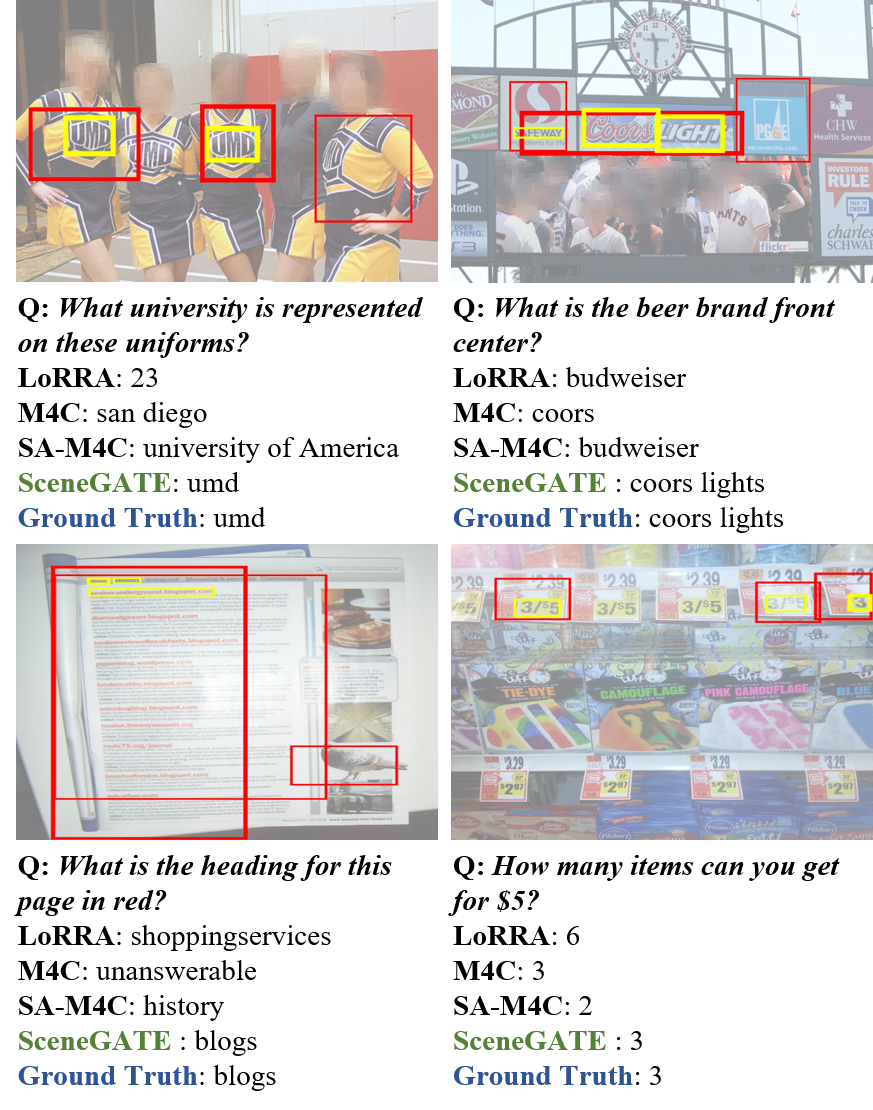}
    \caption{{Visualization} 
 of attention outputs from SceneGATE. Yellow bounding boxes surround the OCR tokens predicted by SceneGATE. Red bounding boxes are the object regions that are associated with the OCR tokens. The thicker the bounding box lines, the higher the attention weights are.}
    \label{fig:SceneGATE_sample}
\end{figure}

Figure~\ref{fig:SceneGATE_sample} shows some sample pairs of images with questions and the answers from different baseline models. The OCR tokens and their associated object regions with high attention weights are highlighted with yellow and red bounding boxes in the images. Compared with other baselines, our model generated more accurate and complete answers with the correct corresponding OCR tokens regions detected in the images. For example, our model perfectly identified the brand of the beer with the answer \textit{Coors Light}, while M4C missed the token \textit{Light} and LoRRA and SA-M4C gave incorrect results for the case in the top-right image. In addition, our model also showed good inference ability in addition to text-reading ability. Taking the bottom-right image as an example, to answer the question of \textit{How many items can you get for \$5?}, the model can not only recognize the correct location of \textit{\$5}, but it also has the ability to understand the semantic meaning of the \emph{forward slash} in the image and to interpret the character before this symbol as a number. Our model provided the correct answer, while the answers of LoRRA and SA-M4C were incorrect. We present more examples and error analyses in the {Appendices \ref{sec:appendix-additional} and \ref{sec:appendix-error}.}

\section{Conclusions}
We propose SceneGATE with the use of a novel TextVQA-based scene graph by treating the OCR tokens in images as the attributes of the objects. Our SceneGATE applies semantic relation-aware attention to the scene graph and uses the guided attention mechanism to obtain the question-guided object and OCR token features, which are then fed into the graph attention module for the learning of the positional relationships between objects and OCR tokens. Our SceneGATE comprehensively learns the semantic and positional relationships between objects and texts in images and outperforms the SOTA on both the Text-VQA and ST-VQA datasets.

\vspace{6pt} 





\authorcontributions{Conceptualization, S.C.H.; investigation, F.C., S.L., F.N. and Z.W.; methodology, F.C. and S.L.; validation, F.C.; formal analysis, S.L.; visualization, F.C.; writing---original draft preparation, F.C. and S.L.; writing---review and editing, S.C.H. and J.P.; supervision, S.C.H.; project administration, S.C.H. All authors have read and agreed to the published version of the manuscript.}

\funding{{This research received no external funding.}}

\institutionalreview{Not applicable.}

\informedconsent{Not applicable.}

\dataavailability{Publicly available datasets were analyzed in this study. These data can be found {at} \url{https://textvqa.org/} for the TextVQA dataset {and} \url{https://rrc.cvc.uab.es/?ch=11&com=introduction} for the ST-VQA dataset. {Both datasets are accessed on 30 June 2023.}}

\conflictsofinterest{The authors declare no conflict of interest.} 






\abbreviations{Abbreviations}{
~~~~~~~~The following abbreviations are used in this manuscript:\\

\noindent 
\begin{tabular}{@{}ll}
VQA & Visual Question Answering\\
OCR & Optical Character Recognition\\
Faster-RCNN & Faster Region-Based Convolutional Neural Network\\
SG & Scene Graph\\
GCN & Graph Convolutional Networks\\
NER & Named Entity Recognition\\
SA & Self-Attention\\
GA & Guided Attention\\
SRA & Semantic Relation-Aware\\
PRA & Positional Relation-Aware\\
MMTE & Multi-Modal Transformer Encoder\\
SOTA & State-of-the-Art\\
VCR & Visual Commonsense Reasoning \\
UMD & University of Maryland \\
ReLU & Rectified Linear unit

\end{tabular}
}


\appendixtitles{yes} 
\appendixstart
\appendix

\section[\appendixname~\thesection]{Training Hyperparameters}\label{appendix-hp}
In this section, we include a list of hyperparameters in Table~\ref{tab:hp}. 

\begin{table}[H]
\caption{{Hyperparameters} used to train the SceneGATE model.}\label{tab:hp}
\setlength{\tabcolsep}{4.9mm} \begin{tabular}{lclc} 
 \toprule
 \textbf{Name} & \textbf{Value} & \textbf{Name} & \textbf{Value}\\
 \midrule
    max text token length & 20 &
    max obj num & 100  \\
    max ocr num & 50 &
    max SG obj num & 36 \\
    max SG ocr num & 100 &
    batch size & 8 \\
    learning rate & 0.0001 &
    \# epoch & 100 \\
    max gradient norm & 0.25 &
    optimizer & Adam \\
    lr decay @ step & 14,000, 19,000 &
    lr decay rate & 0.1 \\
    warmup factor & 0.2 &
    warmup iterations & 1000 \\
    \# workers & 0 &
    distance threshold & 0.5 \\
    SRA Attention context & 3 &
    PRA Attention context & 3 \\
    seed & 0 &
    obj dropout rate & 0.1 \\
    ocr dropout rate & 0.1 &
    hidden size & 768 \\
    \# positional relations & 12 &
    \# semantic relations & 12 \\
    textual query size & 768 &
    ocr feature size & 3002 \\
    obj feature size & 2048 &
    sg feature size & 300 \\
    \# decoding\_steps & 12 &
    text encoder lr scale & 0.1 \\
    \# text encoder layers & 3 &
    \# PRA layers & 2 \\
    \# SRA layers & 2 &
    \# MMTE layers & 2 \\
 \bottomrule
\end{tabular}
\end{table}

\vspace{-12pt}

\begin{figure}[H]
\includegraphics[width=8.5 cm]{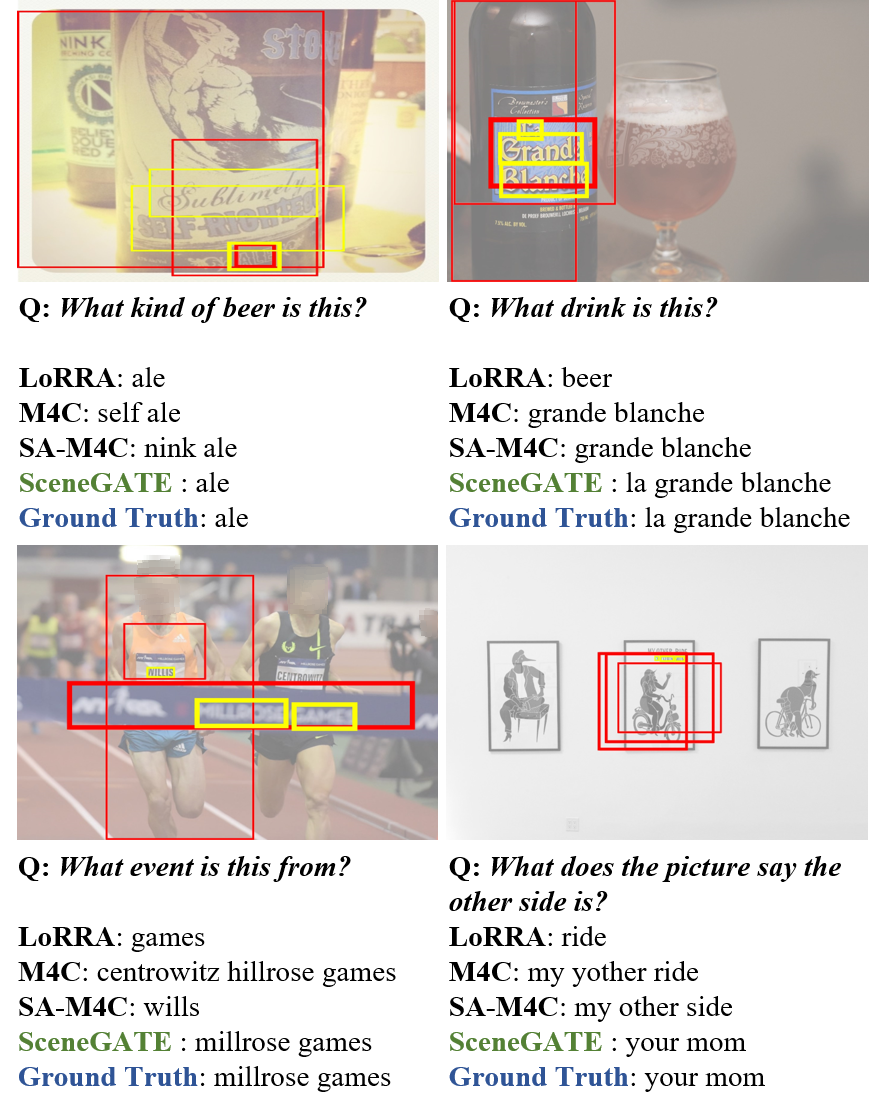}
\caption{Visualization of attention outputs from SceneGATE. Yellow bounding boxes surround the OCR tokens predicted by SceneGATE. Red bounding boxes are the object regions that are associated with the OCR tokens. The thicker the bounding box lines, the higher the attention weights are.\label{fig:SceneGATE_sample_2}}
\end{figure}

\section[\appendixname~\thesection]{Additional Qualitative Examples}\label{sec:appendix-additional}

Figure~\ref{fig:SceneGATE_sample_2} compares some additional prediction results of SceneGATE to those of the other baselines. 

\vspace{-3pt}

\begin{figure}[H]
\includegraphics[width=8 cm]{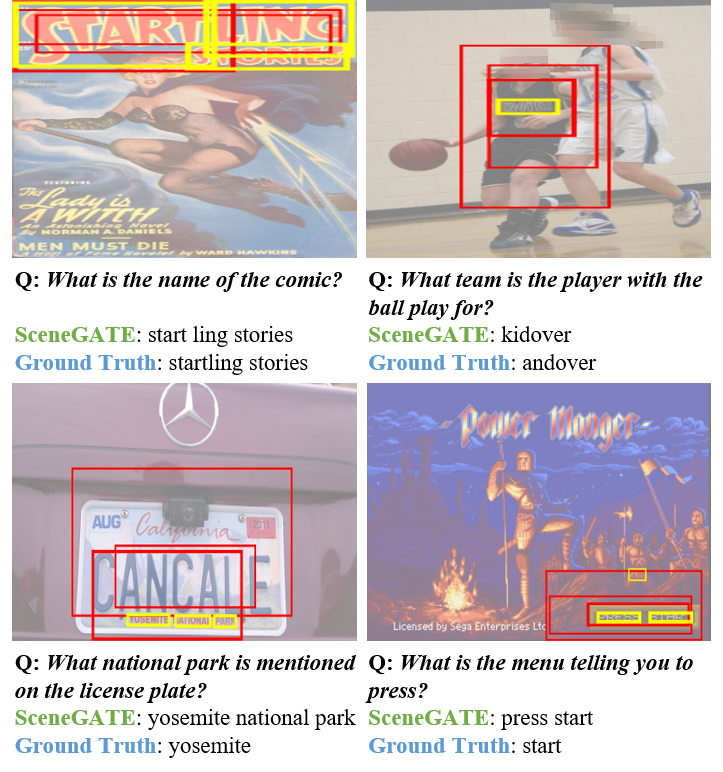}
\caption{Visualization of incorrect classification analysis. Yellow bounding boxes surround the OCR tokens predicted by SceneGATE. Red bounding boxes are the object regions that are associated with the OCR tokens. The thicker the bounding box lines, the higher the attention weights are.\label{fig:error_sample}}
\end{figure}   

\section[\appendixname~\thesection]{Incorrect Classification Analysis}\label{sec:appendix-error}
In Figure~\ref{fig:error_sample}, we show some qualitative examples of the errors that our SceneGATE model makes. 

The first row shows examples of when our model inherits mistakes made by the OCR system. 
Based on the bounding boxes visualized, we can observe that SceneGATE is attending to the correct entities, giving almost correct answers. In the first example, the model is attending to the \textit{startling stories} region of the image, and in the second example, the model attends to the word printed on the black jersey but gives ``kidover'' as the answer instead of ``andover''. This is because the OCR token provided by the pre-trained OCR model is wrong. When the model dynamically copies the OCR tokens as the answer, it inherits the error made by the OCR system despite its own ability to choose the correct elements. Since TextVQA models largely depend on the information of scene texts, the accuracy of pre-trained OCR systems could be a bottleneck for the TextVQA problem. 

The second row shows examples in which our predictions can also be considered as correct answers although they are different from the ground truths. Taking the instance on the left as an example, \textit{yosemite}, \textit{national} and \textit{park} are the top three OCR tokens that our model attends to, and our model outputs all three words as the answer, which is actually correct to answer the given question. Similarly, in the example on the right, the button to be pressed is called \textit{press start}, so our model attends to both words and outputs them as the answer.

\begin{adjustwidth}{-\extralength}{0cm}

\reftitle{References}

\PublishersNote{}
\end{adjustwidth}

\begin{thebibliography}{999}

\bibitem[Singh et~al.(2019)Singh, Natarajan, Shah, Jiang, Chen, Batra, Parikh,
and Rohrbach]{singh2019towards}
Singh, A.; Natarajan, V.; Shah, M.; Jiang, Y.; Chen, X.; Batra, D.; Parikh, D.;
Rohrbach, M.
\newblock Towards vqa models that can read.
\newblock In Proceedings of the IEEE Conference on Computer
Vision and Pattern Recognition, {Long Beach, CA, USA, 15 June 2019;} 
pp.~8317--8326. [\href{http://doi.org/10.1109/CVPR.2019.00851}{CrossRef}]

\bibitem[Hu et~al.(2020)Hu, Singh, Darrell, and Rohrbach]{hu2020iterative}
Hu, R.; Singh, A.; Darrell, T.; Rohrbach, M.
\newblock Iterative answer prediction with pointer-augmented multimodal
transformers for textvqa.
\newblock In Proceedings of the IEEE/CVF Conference on
Computer Vision and Pattern Recognition,  {Seattle, WA, USA, 13--19 June 2020;} pp.~9992--10002. [\href{http://dx.doi.org/10.1109/CVPR42600.2020.01001}{CrossRef}]

\bibitem[Zhu et~al.(2021)Zhu, Gao, Wang, and Wu]{Zhu_Gao_Wang_Wu_2021}
Zhu, Q.; Gao, C.; Wang, P.; Wu, Q.
\newblock Simple is not Easy: A Simple Strong Baseline for TextVQA and
TextCaps.
\newblock {\em Proc. AAAI Conf. Artif. Intell.}
{\bf 2021}, {\em 35},~3608--3615. [\href{http://dx.doi.org/10.1609/aaai.v35i4.16476}{CrossRef}]

\bibitem[Gao et~al.(2020)Gao, Li, Wang, Shan, and Chen]{gao2020multi}
Gao, D.; Li, K.; Wang, R.; Shan, S.; Chen, X.
\newblock Multi-Modal Graph Neural Network for Joint Reasoning on Vision and
Scene Text.
\newblock In Proceedings of the IEEE/CVF Conference on
Computer Vision and Pattern Recognition, {Seattle, WA, USA, 13--19 June 2020;} pp.~12746--12756. [\href{http://dx.doi.org/10.1109/CVPR42600.2020.01276}{CrossRef}]

\bibitem[Kant et~al.(2020)Kant, Batra, Anderson, Schwing, Parikh, Lu, and
Agrawal]{kant2020spatially}
Kant, Y.; Batra, D.; Anderson, P.; Schwing, A.; Parikh, D.; Lu, J.; Agrawal, H.
\newblock Spatially aware multimodal transformers for textvqa.
\newblock In Proceedings of the Computer Vision--ECCV 2020: 16th European
Conference, Glasgow, UK, 23--28 August 2020; Proceedings, Part IX 16;
Springer:  {Berlin/Heidelberg, Germany,} 
2020; pp.~715--732.
\newblock [\href{http://dx.doi.org/10.1007/978-3-030-58545-7_41}{CrossRef}]

\bibitem[Han et~al.(2020)Han, Huang, and Han]{han2020finding}
Han, W.; Huang, H.; Han, T.
\newblock Finding the Evidence: Localization-aware Answer Prediction for Text
Visual Question Answering.
\newblock In Proceedings of the 28th International
Conference on Computational Linguistics,  {Barcelona, Spain, 8--13 December 2020;} pp.~3118--3131. [\href{http://dx.doi.org/10.18653/v1/2020.coling-main.278}{CrossRef}]

\bibitem[Gao et~al.(2022)Gao, Zhu, Wang, Li, Liu, Hengel, and
Wu]{gao2020structured}
Gao, C.; Zhu, Q.; Wang, P.; Li, H.; Liu, Y.; Hengel, A.v.d.; Wu, Q.
\newblock Structured Multimodal Attentions for TextVQA.
\newblock {\em IEEE Trans. Pattern Anal. Mach. Intell.}
{\bf 2022}, {\em 44},~9603--9614. [\href{http://dx.doi.org/10.1109/TPAMI.2021.3132034}{CrossRef}] [\href{http://www.ncbi.nlm.nih.gov/pubmed/34855584}{PubMed}]

\bibitem[Johnson et~al.(2018)Johnson, Gupta, and Fei-Fei]{johnson2018image}
Johnson, J.; Gupta, A.; Fei-Fei, L.
\newblock Image generation from scene graphs.
\newblock In Proceedings of the IEEE conference on computer
vision and pattern recognition, {Salt Lake City, UT, USA, 18--23 June 2018}; pp.~1219--1228. [\href{http://dx.doi.org/10.1109/CVPR.2018.00133}{CrossRef}]

\bibitem[Lu et~al.(2021)Lu, Fan, Wang, Oh, and Rosé]{lu2021localize}
Lu, X.; Fan, Z.; Wang, Y.; Oh, J.; Rosé, C.P.
\newblock Localize, Group, and Select: Boosting Text-VQA by Scene Text
Modeling.
\newblock In Proceedings of the 2021 IEEE/CVF International Conference on
Computer Vision Workshops (ICCVW), {Montreal, BC, Canada, 11--17 October  2021}; pp.~2631--2639. [\href{http://dx.doi.org/10.1109/ICCVW54120.2021.00297}{CrossRef}]

\bibitem[Zeng et~al.(2021)Zeng, Zhang, Zhou, and Yang]{zeng2021beyond}
Zeng, G.; Zhang, Y.; Zhou, Y.; Yang, X. Beyond OCR + VQA: Involving OCR into
the Flow for Robust and Accurate TextVQA.
\newblock In {Proceedings of the 29th ACM International Conference on
Multimedia, Virtual Event China, 20--24 October 2021}; Association for Computing Machinery: New York, NY, USA,  2021; pp. 376–385. [\href{http://dx.doi.org/10.1145/3474085.3475606}{CrossRef}]

\bibitem[Antol et~al.(2015)Antol, Agrawal, Lu, Mitchell, Batra, Zitnick, and
Parikh]{antol2015vqa}
Antol, S.; Agrawal, A.; Lu, J.; Mitchell, M.; Batra, D.; Zitnick, C.L.; Parikh,
D.
\newblock VQA: Visual Question Answering.
\newblock In Proceedings of the 2015 IEEE International Conference on Computer
Vision (ICCV), {Santiago, Chile, 7--13 December 2015}; pp.~2425--2433. [\href{http://dx.doi.org/10.1109/ICCV.2015.279}{CrossRef}]

\bibitem[Shrestha et~al.(2019)Shrestha, Kafle, and Kanan]{shrestha2019answer}
Shrestha, R.; Kafle, K.; Kanan, C.
\newblock Answer them all! toward universal visual question answering models.
\newblock In Proceedings of the Conference on Computer
Vision and Pattern Recognition, {Long Beach, CA, USA, 15--20 June 2019}; pp.~10472--10481. [\href{http://dx.doi.org/10.1109/CVPR.2019.01072}{CrossRef}]

\bibitem[Ben-Younes et~al.(2019)Ben-Younes, Cadene, Thome, and
Cord]{ben2019block}
Ben-Younes, H.; Cadene, R.; Thome, N.; Cord, M.
\newblock Block: Bilinear superdiagonal fusion for visual question answering
and visual relationship detection.
\newblock In Proceedings of the AAAI Conference on
Artificial Intelligence, {Honolulu, HI, USA, 27--28 January 2019}; Volume~33, pp.~8102--8109. [\href{http://dx.doi.org/10.1609/aaai.v33i01.33018102}{CrossRef}]

\bibitem[Cadene et~al.(2019)Cadene, Ben-Younes, Cord, and
Thome]{cadene2019murel}
Cadene, R.; Ben-Younes, H.; Cord, M.; Thome, N.
\newblock Murel: Multimodal relational reasoning for visual question answering.
\newblock In Proceedings of the IEEE/CVF Conference on
Computer Vision and Pattern Recognition,  {Long Beach, CA, USA, 15--20 June 2019}; pp.~1989--1998. [\href{http://dx.doi.org/10.1109/CVPR.2019.00209}{CrossRef}]

\bibitem[Urooj et~al.(2020)Urooj, Mazaheri, Da~vitoria lobo, and
Shah]{urooj2020mmft}
Urooj, A.; Mazaheri, A.; Da~vitoria lobo, N.; Shah, M.
\newblock {MMFT-BERT}: {M}ultimodal {F}usion {T}ransformer with {B}{ERT}
{E}ncodings for {V}isual {Q}uestion {A}nswering.
\newblock In Proceedings of the Findings of the Association for Computational
Linguistics: EMNLP 2020, {Online Event, 16--20 November 2020}; pp.~4648--4660. [\href{http://dx.doi.org/10.18653/v1/2020.findings-emnlp.417}{CrossRef}]

\bibitem[Han et~al.(2021)Han, Guo, Yin, Liu, Hu, and Nie]{han2021focal}
Han, Y.; Guo, Y.; Yin, J.; Liu, M.; Hu, Y.; Nie, L.
\newblock Focal and Composed Vision-Semantic Modeling for Visual Question
Answering.
\newblock In Proceedings of the 29th ACM International
Conference on Multimedia (MM '21), Virtual Event China, 20--24 October 2021; {Association for Computing Machinery: New York, NY, USA,}  2021; pp.~4528–4536.
\newblock [\href{http://dx.doi.org/10.1145/3474085.3475609}{CrossRef}]

\bibitem[Hudson and Manning(2018)]{hudson2018compositional}
Hudson, D.A.; Manning, C.D.
\newblock Compositional Attention Networks for Machine Reasoning.
\newblock In Proceedings of the International Conference on Learning
Representations, {Vancouver, BC, Canada, 30 April--3 May 2018.}

\bibitem[Anderson et~al.(2018)Anderson, He, Buehler, Teney, Johnson, Gould, and
Zhang]{anderson2018bottom}
Anderson, P.; He, X.; Buehler, C.; Teney, D.; Johnson, M.; Gould, S.; Zhang, L.
\newblock Bottom-up and top-down attention for image captioning and visual
question answering.
\newblock In Proceedings of the IEEE conference on computer
vision and pattern recognition, {Salt Lake City, UT, USA, 18--23 June 2018}; pp.~6077--6086. [\href{http://dx.doi.org/10.1109/CVPR.2018.00636}{CrossRef}]

\bibitem[Yu et~al.(2019)Yu, Yu, Cui, Tao, and Tian]{yu2019deep}
Yu, Z.; Yu, J.; Cui, Y.; Tao, D.; Tian, Q.
\newblock Deep modular co-attention networks for visual question answering.
\newblock In Proceedings of the IEEE conference on computer
vision and pattern recognition, {Long Beach, CA, USA, 15--20 June 2019}; pp.~6281--6290. [\href{http://dx.doi.org/10.1109/CVPR.2019.00644}{CrossRef}]

\bibitem[Gao et~al.(2019)Gao, You, Zhang, Wang, and Li]{gao2019multi}
Gao, P.; You, H.; Zhang, Z.; Wang, X.; Li, H.
\newblock Multi-modality latent interaction network for visual question
answering.
\newblock In Proceedings of the IEEE/CVF International
Conference on Computer Vision, {Seoul, Republic of Korea, 27 Octorber--2 November 2019;} pp.~5825--5835. [\href{http://dx.doi.org/10.1109/ICCV.2019.00592}{CrossRef}]

\bibitem[Nguyen and Okatani(2018)]{nguyen2018improved}
Nguyen, D.K.; Okatani, T.
\newblock Improved fusion of visual and language representations by dense
symmetric co-attention for visual question answering.
\newblock In Proceedings of the IEEE Conference on Computer Vision and Pattern
Recognition, {Salt Lake City, UT, USA, 18--23 June 2018}; pp.~6087--6096. [\href{http://dx.doi.org/10.1109/CVPR.2018.00637}{CrossRef}]

\bibitem[Rahman et~al.(2021)Rahman, Chou, Sigal, and
Carenini]{rahman2021improved}
Rahman, T.; Chou, S.H.; Sigal, L.; Carenini, G.
\newblock An Improved Attention for Visual Question Answering.
\newblock In Proceedings of the 2021 IEEE/CVF Conference on Computer Vision and
Pattern Recognition Workshops (CVPRW), {Nashville, TN, USA, 19--25 June 2021}; pp.~1653--1662. [\href{http://dx.doi.org/10.1109/CVPRW53098.2021.00181}{CrossRef}]

\bibitem[Yang et~al.(2019)Yang, Tang, Zhang, and Cai]{yang2019auto}
Yang, X.; Tang, K.; Zhang, H.; Cai, J.
\newblock Auto-encoding scene graphs for image captioning.
\newblock In Proceedings of the IEEE/CVF Conference on
Computer Vision and Pattern Recognition, {Long Beach, CA, USA, 15--20 June 2019}; pp.~10685--10694. [\href{http://dx.doi.org/10.1109/CVPR.2019.01094}{CrossRef}]

\bibitem[Gu et~al.(2019)Gu, Joty, Cai, Zhao, Yang, and Wang]{gu2019unpaired}
Gu, J.; Joty, S.; Cai, J.; Zhao, H.; Yang, X.; Wang, G.
\newblock Unpaired image captioning via scene graph alignments.
\newblock In Proceedings of the IEEE/CVF International
Conference on Computer Vision,  {Long Beach, CA, USA, 15--20 June 2019}; pp.~10323--10332. [\href{http://dx.doi.org/10.1109/ICCV.2019.01042}{CrossRef}]

\bibitem[Han et~al.(2020)Han, Long, Luo, Wang, and Poon]{han2020victr}
Han, C.; Long, S.; Luo, S.; Wang, K.; Poon, J.
\newblock VICTR: Visual Information Captured Text Representation for
Text-to-Vision Multimodal Tasks.
\newblock In Proceedings of the 28th International
Conference on Computational Linguistics, {Barcelona, Spain, 8--13 December 2020}; pp.~3107--3117. [\href{http://dx.doi.org/10.18653/v1/2020.coling-main.277}{CrossRef}]

\bibitem[Wang et~al.(2020)Wang, Wang, Yao, Shan, and Chen]{wang2020cross}
Wang, S.; Wang, R.; Yao, Z.; Shan, S.; Chen, X.
\newblock Cross-modal scene graph matching for relationship-aware image-text
retrieval.
\newblock In Proceedings of the IEEE/CVF Winter Conference
on Applications of Computer Vision, {Snowmass Village, CO, USA, 1--5 March 2020}; pp.~1508--1517. [\href{http://dx.doi.org/10.1109/WACV45572.2020.9093614}{CrossRef}]

\bibitem[Luo et~al.(2020)Luo, Han, Sun, and Poon]{luo2020rexup}
Luo, S.; Han, S.C.; Sun, K.; Poon, J.
\newblock REXUP: I REason, I EXtract, I UPdate with Structured Compositional
Reasoning for Visual Question Answering.
\newblock In Proceedings of the International Conference on Neural Information
Processing, {Bangkok, Thailand, 18--22 November 2020}; Springer:  {Berlin/Heidelberg, Germany,} 
2020; pp.~520--532.
\newblock [\href{http://dx.doi.org/10.1007/978-3-030-63830-6_44}{CrossRef}]

\bibitem[Hudson and Manning(2019)]{hudson2019learning}
Hudson, D.; Manning, C.D.
\newblock Learning by abstraction: The neural state machine.
\newblock In Proceedings of the Advances in Neural Information Processing
Systems, {Vancouver, BC, Canada, 8--14 December 2019}; pp.~5903--5916.

\bibitem[Haurilet et~al.(2019)Haurilet, Roitberg, and
Stiefelhagen]{haurilet2019s}
Haurilet, M.; Roitberg, A.; Stiefelhagen, R.
\newblock It's Not About the Journey; It's About the Destination: Following
Soft Paths Under Question-Guidance for Visual Reasoning.
\newblock In Proceedings of the IEEE Conference on Computer
Vision and Pattern Recognition, {Long Beach, CA, USA, 15--20 June 2019}; pp.~1930--1939. [\href{http://dx.doi.org/10.1109/CVPR.2019.00203}{CrossRef}]

\bibitem[Nuthalapati et~al.(2021)Nuthalapati, Chandradevan, Giunchiglia, Li,
Kayser, Lukasiewicz, and Yang]{nuthalapati2021lightweight}
Nuthalapati, S.V.; Chandradevan, R.; Giunchiglia, E.; Li, B.; Kayser, M.;
Lukasiewicz, T.; Yang, C. Lightweight Visual Question Answering Using Scene
Graphs.
\newblock In Proceedings of the 30th ACM International Conference on
Information \& Knowledge Management, {Virtual Event, 1--5 November 2021}; Association for Computing Machinery: New York, NY, USA,  2021; pp.~3353–3357. [\href{http://dx.doi.org/10.1145/3459637.3482218}{CrossRef}]

\bibitem[Wang et~al.(2022)Wang, You, Li, Zareian, Park, Liang, Chang, and
Chang]{wang2022sgeitl}
Wang, Z.; You, H.; Li, L.H.; Zareian, A.; Park, S.; Liang, Y.; Chang, K.W.;
Chang, S.F.
\newblock SGEITL: Scene Graph Enhanced Image-Text Learning for Visual
Commonsense Reasoning.
\newblock {\em Proc. AAAI Conf. Artif. Intell.}
{\bf 2022}, {\em 36},~5914--5922. [\href{http://dx.doi.org/10.1609/aaai.v36i5.20536}{CrossRef}]

\bibitem[Devlin et~al.(2019)Devlin, Chang, Lee, and
Toutanova]{devlin-etal-2019-bert}
Devlin, J.; Chang, M.W.; Lee, K.; Toutanova, K.
\newblock {BERT}: Pre-training of Deep Bidirectional Transformers for Language
Understanding.
\newblock In Proceedings of the 2019 Conference of the North
{A}merican Chapter of the Association for Computational Linguistic (NAACL),
Minneapolis, MN, USA, 2--7 June 2019; pp.~4171--4186. [\href{http://dx.doi.org/10.18653/v1/N19-1423}{CrossRef}]

\bibitem[Krishna et~al.(2017)Krishna, Zhu, Groth, Johnson, Hata, Kravitz, Chen,
Kalantidis, Li, Shamma, Bernstein, and Fei-Fei]{Krishna2017VG}
Krishna, R.; Zhu, Y.; Groth, O.; Johnson, J.; Hata, K.; Kravitz, J.; Chen, S.;
Kalantidis, Y.; Li, L.J.; Shamma, D.A.;  et~al.
\newblock Visual Genome: Connecting Language and Vision Using Crowdsourced
Dense Image Annotations.
\newblock {\em Int. J. Comput. Vis.} {\bf 2017}, {\em 123},~32–73. [\href{http://dx.doi.org/10.1007/s11263-016-0981-7}{CrossRef}]

\bibitem[Almazán et~al.(2014)Almazán, Gordo, Fornés, and Valveny]{phoc}
Almazán, J.; Gordo, A.; Fornés, A.; Valveny, E.
\newblock Word Spotting and Recognition with Embedded Attributes.
\newblock {\em IEEE Trans. Pattern Anal. Mach. Intell.}
{\bf 2014}, {\em 36},~2552--2566. [\href{http://dx.doi.org/10.1109/TPAMI.2014.2339814}{CrossRef}] [\href{http://www.ncbi.nlm.nih.gov/pubmed/26353157}{PubMed}]
\clearpage
\bibitem[Vaswani et~al.(2017)Vaswani, Shazeer, Parmar, Uszkoreit, Jones, Gomez,
Kaiser, and Polosukhin]{vaswani2017tranformer}
Vaswani, A.; Shazeer, N.; Parmar, N.; Uszkoreit, J.; Jones, L.; Gomez, A.N.;
Kaiser, L.u.; Polosukhin, I.
\newblock Attention is All you Need.
\newblock In Proceedings of the Advances in Neural Information Processing
{Systems, Long Beach, CA, USA, 4--9 December 2017;} Guyon, I., Luxburg, U.V., Bengio, S., Wallach, H., Fergus, R.,
Vishwanathan, S., Garnett, R., Eds.; {Curran Associates, Inc.: Red Hook, NY, USA}, 
2017; Volume~30.

\bibitem[Krasin et~al.(2017)Krasin, Duerig, Alldrin, Ferrari, Abu-El-Haija,
Kuznetsova, Rom, Uijlings, Popov, Veit, Belongie, Gomes, Gupta, Sun, Chechik,
Cai, Feng, Narayanan, and Murphy]{openimages}
Krasin, I.; Duerig, T.; Alldrin, N.; Ferrari, V.; Abu-El-Haija, S.; Kuznetsova,
A.; Rom, H.; Uijlings, J.; Popov, S.; Veit, A.;  et~al.
\newblock OpenImages: A public dataset for large-scale multi-label and
multi-class image classification.  {\bf 2017}, \emph{2}, 18.
\newblock Available online: \url{https://github.com/openimages} {(accessed on 30 June 2023).} 
\bibitem[Biten et~al.(2019)Biten, Tito, Mafla, Gomez, Rusinol, Valveny,
Jawahar, and Karatzas]{biten2019scene}
Biten, A.F.; Tito, R.; Mafla, A.; Gomez, L.; Rusinol, M.; Valveny, E.; Jawahar,
C.; Karatzas, D.
\newblock Scene text visual question answering.
\newblock In Proceedings of the IEEE/CVF International
Conference on Computer Vision, {Seoul, Republic of Korea, 27 Octorber--2 November 2019;} pp.~4291--4301. [\href{http://dx.doi.org/10.1109/ICCV.2019.00439}{CrossRef}]

\bibitem[Veit et~al.(2016)Veit, Matera, Neumann, Matas, and
Belongie]{veit2016cocotext}
Veit, A.; Matera, T.; Neumann, L.; Matas, J.; Belongie, S.
\newblock COCO-Text: Dataset and Benchmark for Text Detection and Recognition
in Natural Images.
\newblock {\em arXiv} {\bf 2016}, arXiv:1601.07140.

\bibitem[Gurari et~al.(2018)Gurari, Li, Stangl, Guo, Lin, Grauman, Luo, and
Bigham]{gurari2018vizwiz}
Gurari, D.; Li, Q.; Stangl, A.J.; Guo, A.; Lin, C.; Grauman, K.; Luo, J.;
Bigham, J.P.
\newblock VizWiz Grand Challenge: Answering Visual Questions from Blind People.
\newblock In Proceedings of the 2018 IEEE/CVF Conference on Computer Vision and
Pattern Recognition, {Salt Lake City, UT, USA, 18--23 June 2018}; pp.~3608--3617. [\href{http://dx.doi.org/10.1109/CVPR.2018.00380}{CrossRef}]

\bibitem[Karatzas et~al.(2013)Karatzas, Shafait, Uchida, Iwamura, Bigorda,
Mestre, Mas, Mota, Almaz\`{a}n, and de~las Heras]{Karatzas2013ICDAR}
Karatzas, D.; Shafait, F.; Uchida, S.; Iwamura, M.; Bigorda, L.G.i.; Mestre,
S.R.; Mas, J.; Mota, D.F.; Almaz\`{a}n, J.A.; de~las Heras, L.P.
\newblock ICDAR 2013 Robust Reading Competition.
\newblock In Proceedings of the 2013 12th International
Conference on Document Analysis and Recognition (ICDAR '13),   Washington, DC, USA, 25–28 August 2013;
pp.~1484–1493. [\href{http://dx.doi.org/10.1109/ICDAR.2013.221}{CrossRef}]

\bibitem[Karatzas et~al.(2015)Karatzas, Gomez-Bigorda, Nicolaou, Ghosh,
Bagdanov, Iwamura, Matas, Neumann, Chandrasekhar, Lu, Shafait, Uchida, and
Valveny]{Karatzas2015ICDAR}
Karatzas, D.; Gomez-Bigorda, L.; Nicolaou, A.; Ghosh, S.; Bagdanov, A.;
Iwamura, M.; Matas, J.; Neumann, L.; Chandrasekhar, V.R.; Lu, S.;  et~al.
\newblock ICDAR 2015 competition on Robust Reading.
\newblock In Proceedings of the 2015 13th International Conference on Document
Analysis and Recognition (ICDAR), {Tunis, Tunisia, 23--26 August 2015}; pp.~1156--1160. [\href{http://dx.doi.org/10.1109/ICDAR.2015.7333942}{CrossRef}]

\bibitem[Deng et~al.(2009)Deng, Dong, Socher, Li, Li, and
Fei-Fei]{deng2009imagenet}
Deng, J.; Dong, W.; Socher, R.; Li, L.J.; Li, K.; Fei-Fei, L.
\newblock ImageNet: A large-scale hierarchical image database.
\newblock In Proceedings of the 2009 IEEE Conference on Computer Vision and
Pattern Recognition, {Miami, FL, USA, 20--25 June 2009}; pp.~248--255. [\href{http://dx.doi.org/10.1109/CVPR.2009.5206848}{CrossRef}]

\bibitem[Mishra et~al.(2013)Mishra, Alahari, and Jawahar]{mishra2013iiitstr}
Mishra, A.; Alahari, K.; Jawahar, C.
\newblock Image Retrieval Using Textual Cues.
\newblock In Proceedings of the 2013 IEEE International Conference on Computer
Vision, {Sydney, Australia, 1--8 December 2013}; pp.~3040--3047. [\href{http://dx.doi.org/10.1109/ICCV.2013.378}{CrossRef}]


\bibitem[Goyal et~al.(2017)Goyal, Khot, Summers-Stay, Batra, and
Parikh]{goyal2017making}
Goyal, Y.; Khot, T.; Summers-Stay, D.; Batra, D.; Parikh, D.
\newblock Making the V in VQA Matter: Elevating the Role of Image Understanding
in Visual Question Answering.
\newblock In Proceedings of the 2017 IEEE Conference on Computer Vision and
Pattern Recognition (CVPR), {Honolulu, HI, USA, 21--26 July 2017}; pp.~6325--6334. [\href{http://dx.doi.org/10.1109/CVPR.2017.670}{CrossRef}]

\bibitem[Johnson et~al.(2017)Johnson, Hariharan, van~der Maaten, Fei-Fei,
Zitnick, and Girshick]{johnson2017clevr}
Johnson, J.; Hariharan, B.; van~der Maaten, L.; Fei-Fei, L.; Zitnick, C.L.;
Girshick, R.
\newblock CLEVR: A Diagnostic Dataset for Compositional Language and Elementary
Visual Reasoning.
\newblock In Proceedings of the 2017 IEEE Conference on Computer Vision and
Pattern Recognition (CVPR), {Honolulu, HI, USA, 21--26 July 2017}; pp.~1988--1997. [\href{http://dx.doi.org/10.1109/CVPR.2017.215}{CrossRef}]

\bibitem[Hudson and Manning(2019)]{hudson2019gqa}
Hudson, D.A.; Manning, C.D.
\newblock GQA: A New Dataset for Real-World Visual Reasoning and Compositional
Question Answering.
\newblock In Proceedings of the 2019 IEEE/CVF Conference on Computer Vision and
Pattern Recognition (CVPR), {Long Beach, CA, USA, 15--20 June 2019}; pp.~6693--6702. [\href{http://dx.doi.org/10.1109/CVPR.2019.00686}{CrossRef}]





\end{thebibliography}
\end{document}